\documentclass[11pt,a4paper]{article}
\usepackage{times}
\usepackage{latexsym}
\usepackage{url}
\usepackage[T1]{fontenc}

\usepackage[acceptedWithA]{tacl2021v1}

\usepackage{xspace,mfirstuc,tabulary}

\newif\iftaclinstructions
\taclinstructionsfalse %
\iftaclinstructions
\renewcommand{\confidential}{}
\renewcommand{\anonsubtext}{(No author info supplied here, for consistency with
TACL-submission anonymization requirements)}
\newcommand{\instr}
\fi

\iftaclpubformat %

\else

\fi

\usepackage{microtype}

\usepackage{amsmath, amssymb}
\usepackage{inconsolata}
\usepackage{color}
\usepackage{tikz}
\usepackage{colortbl} %
\usepackage{framed}
\usepackage{multirow}
\usepackage{hyperref}
\usepackage{flushend}
\usepackage{balance}
\usepackage{url}
\usepackage{verbatim}
\usepackage{xspace}
\usepackage{arydshln}
\usepackage{enumitem}
\usepackage{graphicx}
\usepackage{booktabs}
\usepackage{xcolor}
\usepackage{soul}
\usepackage{nicematrix}
\usepackage[multiple]{footmisc}

\usepackage[ruled,vlined]{algorithm2e}
\SetKwInput{KwInitialize}{Initialize} %
\SetKwComment{tcp}{\textcolor{blue}{// }}{}
\SetKwComment{Comment}{}{}

\newcommand{\squishlist}{
 \begin{list}{$\bullet$}
  { \setlength{\itemsep}{0pt}
     \setlength{\parsep}{1pt}
     \setlength{\topsep}{1pt}
     \setlength{\partopsep}{0pt}
     \setlength{\leftmargin}{1.5em}
     \setlength{\labelwidth}{1em}
     \setlength{\labelsep}{0.5em} } }
 \newcommand{\squishend}{\end{list}}

\usepackage{array}
\newcolumntype{?}{!{\vrule width 2pt}}

\newcolumntype{Q}[1]{>{\centering\arraybackslash}p{#1}}
\newcolumntype{P}[1]{>{\centering\arraybackslash}p{#1}}

\author{Sneha Singhania \\
MPI for Informatics\\
\normalsize{ssinghan@mpi-inf.mpg.de} \\\And
Simon Razniewski\\
ScaDS.AI \& TUD \\
\normalsize{simon.razniewski@tu-dresden.de} \\\And
Gerhard Weikum\\
MPI for Informatics\\
\normalsize{weikum@mpi-inf.mpg.de} \\}

\title{Recall Them All: Retrieval-Augmented Language Models for\\ Long Object List Extraction from Long Documents}

\begin{document}
\maketitle

\begin{abstract}
Relation extraction methods from text often prioritize high precision but at the expense of recall. However, high recall is crucial for populating long lists of object entities that stand in a specific relation with a given subject. In long texts, cues for relevant objects can be spread across many passages, posing a challenge for extracting long lists. We present the L3X method which tackles this problem in two stages: (1) recall-oriented generation using a large language model with judicious techniques for retrieval augmentation,  and (2) precision-oriented scrutinization to validate or prune candidates.
\end{abstract}

\section{Introduction}

\textbf{Motivation and Problem.}
Information extraction (IE, for short) is the methodology for distilling structured information out of unstructured texts. Specifically, relation extraction aims to yield subject-predicate-object ({\em SPO}) triples where S and O are named entities that stand in a certain relation P.
State-of-the-art methods are based on neural learning, and perform well in terms of precision but with limited recall~\cite{HanGLPYXLLZS20}. Recently, large language models (LLM) have been studied for this task, with emphasis on long-tail facts, yet they exhibit similar deficits in recall ~\cite{DBLP:conf/icml/KandpalDRWR23, DBLP:conf/emnlp/VeseliRKW23, DBLP:journals/corr/abs-2308-10168}. Moreover, most methods are designed to operate only on single passages, as classifiers or sequence taggers.

\begin{figure}[t]
  \includegraphics[width=0.9\columnwidth]{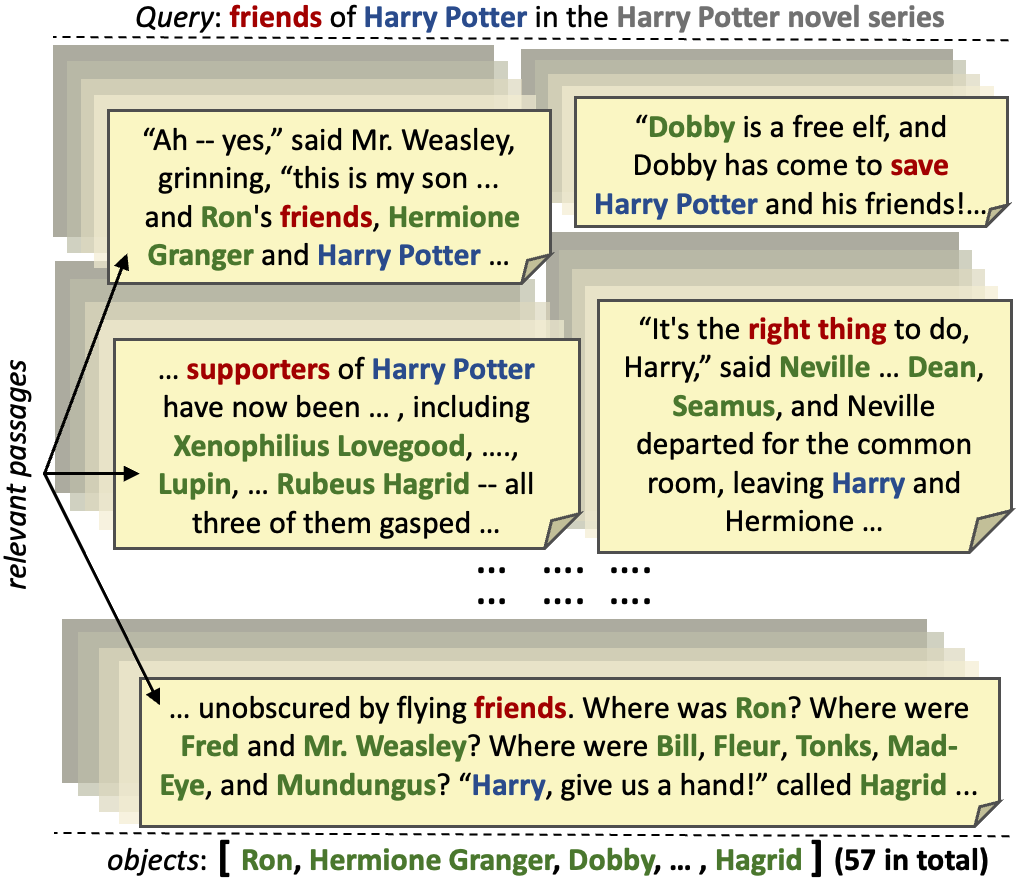}
  \caption{Example for extracting long lists from long texts. For the subject ``Harry Potter'', we aim to extract all 57 friends, appearing throughout the book series.}
  \label{fig:problem-illustration}
\end{figure}

This work addresses the underexplored and unsolved problem that IE faces with two \textit{``longs''}: extracting a \textbf{long list} of object entities that stand in a certain relation to a subject, appearing in \textbf{long text}, such as entire books or websites with many pages. Examples for this open challenge would be extracting a complete list of (nearly) all acquisitions and subsidiaries of Alphabet Inc., identifying all artists who have covered Bob Dylan songs, or finding all friends of Harry Potter in the Harry Potter book series. Figure \ref{fig:problem-illustration} illustrates this task. 

The most challenging cases arise when a relation is expressed merely by soft cues in a few passages, scattered across an entire book. For instance, consider Harry's friend Neville: Figure \ref{fig:problem-illustration} shows a weak cue---very unwieldy for established IE methods. Additional similar weak cues are spread throughout the books, and only by aggregating all of them can we confidently infer this friendship. 
This calls for novel methods regarding (i) retrieving informative passages, and
(ii) performing inference over multiple passages.

\vspace{0.2cm}
\noindent
\textbf{Approach and Contributions.}
We devise a novel methodology to address this 
challenging task. Our method, called {\bf L3X} (\textbf{L}M-based \textbf{L}ong \textbf{L}ist e\textbf{X}traction), operates in two stages:

\noindent {\bf Stage 1: Recall-oriented Generation}. An LLM is prompted with the subject and relation at hand, and tasked to generate a full list of objects through various prompt formulations. 
In addition, we use information retrieval (IR) methods to find promising candidate passages from long texts and feed them into the LLM prompts.
In contrast to prior works on retrieval-augmented LLMs, we retrieve a large number of such passages (e.g., 500 for a given SP pair) and judiciously select the best ones for prompting. Moreover, our method iteratively re-ranks the passages and re-prompts the LLM, to improve recall of initial generation of objects.
    
\noindent {\bf Stage 2: Precision-oriented Scrutinization}. Given a high-recall list of object candidates from Stage 1, the second stage uses conservative techniques to corroborate or prune objects. We employ novel techniques to identify high-confidence objects and their best support passages, and use them to re-assess lower-confidence candidates. 

Since we are solving a new task, we curated two datasets, covering fiction books and web documents, respectively. The books dataset, which is our primary target, consists of 11 books or book series, with a total of 16,000 pages. It addresses 8 relations of long-tailed nature (incl. friends, opponents, placeHasPerson etc.).
The second dataset comprises ca. 10 million web documents sampled from the C4 corpus~\cite{dodge-etal-2021-documenting}, focusing on 3 long-tailed factual relations (hasCEO, hasSubsidiary, and isMemberOf). 
Here, for each SP pair, we need to tap into many thousands of pages, which can be conceptualized as a single long text.

Due to the inherent trade-off between precision and recall, neither metric alone is suitable for our task, and F1 would merely be a generic compromise.
The task instead requires maximizing recall, for an effect on knowledge graph (KG) population, with sufficiently high precision to keep downstream curation efforts manageable. 
Therefore, the metric that we aim to optimize
is {\bf Recall@PrecisionX (R@P$\mathtt{x}$)}, where $\mathtt{x}$ is the minimum precision target
to be achieved
(e.g., $\mathtt{x}$ being
50\% or, ideally, 80\%).
In experiments with Llama3.1-instruct-70B %
\cite{DBLP:journals/corr/abs-2407-21783}
as underlying LLM, we reach 80-85\% recall using our passage re-ranking and batching technique and ca. 50\% R@P50 and 37\% R@P80 through our scrutinization process.

The salient contributions of this work are:
(1) the new task of extracting a long list of objects for a given subject and 
relation from long documents;
(2) a methodology for this task, based on retrieval-augmented LLMs
and combining IR techniques with LLM generation; 
(3) experiments with new benchmarks, showing that L3X outperforms LLM-only baselines that rely on parametric memory from their pre-training, and providing an in-depth analysis of 
strengths and limitations of different methods.
The new datasets, code, and additional experimental results, will be open-sourced upon publication.

\section{Related Work}

\paragraph{\textbf{Relation Extraction.}}
A common task in IE is to extract the relation P that holds between two given entities,
subject (S) and object (O), where P comes from a 
pre-specified set of possible predicates.
State-of-the-art methods (e.g., \cite{HanGLPYXLLZS20, DBLP:conf/emnlp/WangHCS20, DBLP:conf/emnlp/CabotN21, DBLP:conf/acl/XieSLM022, josifoski-etal-2022-genie, ma-etal-2023-dreeam}) 
typically operate 
on single passages, as input to a multi-label classifier or sequence tagger.

Recent works
~\cite{DBLP:journals/csur/ZhaoDYWZCLSX24, DBLP:journals/fcsc/XuCPZXZWZWC24}
have advanced the scope of the extractors' inputs under the theme of
``long-distance IE'',
going beyond single sentences/passages.
However, techniques like 
graph neural networks or LLM-powered generative IE 
are geared for short news or chats, 
and
cannot cope with
book-length texts.
In the popular document-level benchmark DocRED~\cite{DBLP:conf/acl/YaoYLHLLLHZS19},
inputs
are single paragraphs from Wikipedia.
Aggregating cues from many passages 
(as required, e.g., for determining that Neville is Harry's friend) is out of scope.
Moreover, these prior works assume texts for extraction are given upfront, or retrieved by matching S and O in proximity. 
In contrast, our long-list task takes S and P as input and seeks to generate previously unseen
O as output. 
This changes the goal from high-precision classification to
high-recall extraction.

\paragraph{\textbf{OpenIE.}}
OpenIE~\cite{DBLP:conf/ijcai/Mausam16,DBLP:conf/naacl/StanovskyMZD18,DBLP:conf/acl/KolluruMMCM22} is a variant where S, P and O are simply surface phrases without linkage to a knowledge base. 
While OpenIE may provide broader coverage across different relations, it is unsuitable for populating lists of crisp object entities for a given relation.
Even when powered by distant supervision with (S,O) pairs, it remains limited to extraction from single sentences or short passages
(e.g., 
\cite{DBLP:journals/csur/SmirnovaC19}).

\vspace*{-0.1cm}
\paragraph{\textbf{LLMs as Knowledge Bases.}}
\citet{petroni-etal-2019-language} showed that 
LLM prompts can generate facts of knowledge-base style. 
The approach has been expanded
and refined in various ways (e.g., \cite{DBLP:journals/tacl/JiangXAN20,shin-etal-2020-autoprompt,DBLP:conf/naacl/QinE21,DBLP:conf/www/ChenZXDYTHSC22}). 
These aim at precision, 
disregarding recall and the long tail.
Recent studies indicate that LLMs have major problems in dealing with long tail facts~\cite{DBLP:conf/emnlp/VeseliRKW23,DBLP:journals/corr/abs-2308-10168,singhania-etal-2023-extracting,DBLP:conf/icml/KandpalDRWR23}.

\paragraph{\textbf{Retrieval-Augmented Generation.}}
For better LLM generations, relevant text snippets can be retrieved and fed into in-context prompts, through the popularly known RAG paradigm
\cite{NEURIPS2020_6b493230, realm}. 
The surveys \cite{DBLP:conf/sigir/0002WL022,asai-etal-2023-retrieval-tutorial,DBLP:journals/corr/abs-2310-07521,DBLP:journals/corr/abs-2312-10997}
discuss RAG architectures for improving overall task accuracy.

\paragraph{\textbf{Evidence and Factuality.}}
LLMs can be harnessed 
to assess the factuality of statements (e.g., \cite{DBLP:conf/emnlp/ManakulLG23,DBLP:conf/emnlp/MinKLLYKIZH23,DBLP:journals/corr/abs-2307-13528,DBLP:journals/corr/abs-2310-07521}). 
These techniques leverage external sources, such as Wikipedia articles, which is infeasible in our setting, where the focus is on long-tail entities within long (fictional) books.

\paragraph{\textbf{IE from Books.}} 
Prior works by \citet{bamman-etal-2019-annotated, stammbach-etal-2022-heroes, chang-etal-2023-speak} pursue LLM-supported IE about characters from fiction books. However, these methods focus on generating a single name from a single passage.

\begin{figure*}[tp]
    \centering
    \includegraphics[scale = 0.4]{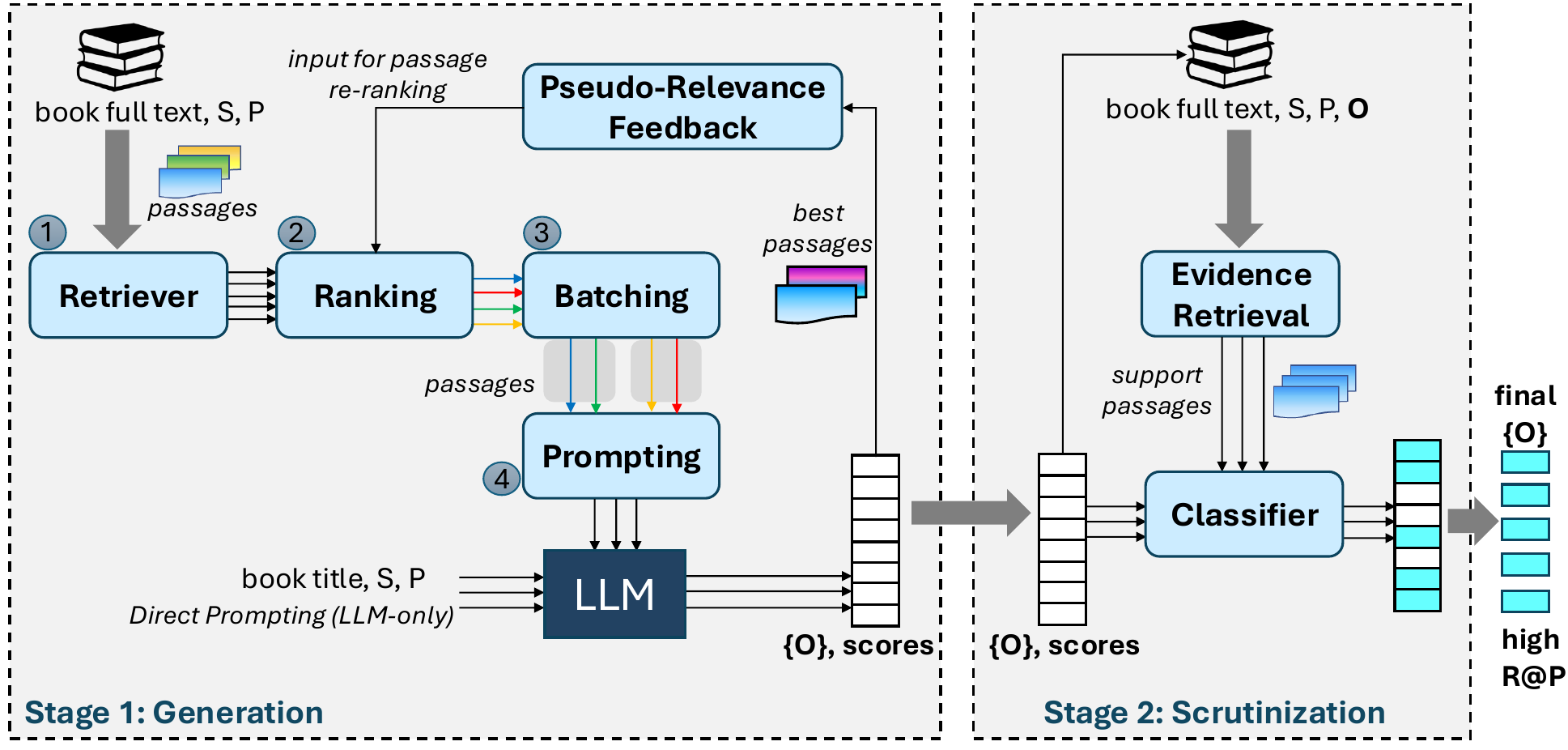}
    \vspace{-0.3cm}
    \caption{Overview of the \textsc{L3X} methodology. 
    }
    \label{fig:methodology}
    \end{figure*}

    \section{Recall-oriented Generation}
    \label{sec:recall}
    
    Our first, recall-oriented stage comprises several steps; Figure~\ref{fig:methodology} (left side)
    provides an overview of the flow between the components.
    
    \squishlist
    \item[1.] {\bf Retrieval} of a large pool of passages from the long text, using a  dense retriever by searching with S and a set of paraphrases of P.
    
    \item[2.]  {\bf Re-ranking} passages by various criteria. We present two techniques to prioritize passages based on (i) {\bf num:} number of {\em named-entity mentions} in a passage, to leverage co-occurrences of multiple O values for the same predicate
    (e.g., a passage about several friends), and (ii) {\bf amp:} pseudo-relevance feedback~\cite{DBLP:journals/ftir/Zhai08} to {\em amplify} signals from best passages to refine the prompt for the next round.
    
    \item[3.] {\bf Batching} passages with (i) {\bf neo:} similar entities (including their aliases) 
    identified through named entity overlap, and (ii) {\bf sim:} similar narratives via embeddings, to provide semantically coherent inputs to the LLM.
    
    \item[4.] {\bf Prompting} the LLM in {\em retrieval-augmented mode} using the top-ranked passages. The prompt explicitly includes the book title, S and P. For recall, this is an {\em ensemble} over different choices of retrieved 
    passage (step 1.), 
    and the output of this stage is the union of all objects generated by the LLM. 
    
    \squishend

    \noindent 
    The first three steps are optional, enabling simpler configurations. Running only step 4 (without retrieved passages) results in an LLM-only/no-RAG variant, serving as a {\bf direct prompting} baseline. Running only steps 1 and 4 produces a simplified variant of L3X, referred to as {\bf def} (for default configuration), where passages are ranked by retriever scores and batched in the same order. Each of these steps is elaborated below.
    As Figure \ref{fig:methodology} shows, some of the steps can also be iterated;
    Section \ref{sec:prf} discusses this for the {\em amp} technique.

    \subsection{Passage Retrieval}\label{subsec:retriever}
    Long texts, like entire books, are chunked into short 
    passages of 15 sentences, totaling up to 1000 characters. 
    We construct all overlapping passages (i.e., sharing sentences), to
    ensure that sentences with co-references stay implicitly connected to named entities in their proximity. 
    Since books contain long pieces of direct speech, which may not mention all speakers' names explicitly, we further enrich each passage with  {\em mentions of people and locations} from the {\em preceding}
    10 passages
    by default. This metadata annotation ensures that useful named entity information from prior chunks is available within the current passage.
    All person-person relations and the placeHasPerson relation can potentially benefit from this form of contextualization.

    On the large pool of 
    enriched
    passages, indexed for efficient retrieval, we experimented with several retrieval models,
    including BM25~\cite{bm25}, Contriever~\cite{contriever}, and techniques based on OpenAI embeddings. Among the best-performing, we selected
    Contriever, a BERT-based dense neural IR method fine-tuned on MS-MARCO~\cite{contriever}, with the added benefit of being open-source\footnote{\url{https://github.com/facebookresearch/contriever}} and deployable under our full control. For brevity, we report only experiments with Contriever.

    \subsection{Passage Ranking}\label{sec:prf}
    The default passage ranking is directly derived from retriever scores.
    Additionally, we propose two
    {\em re-ranking heuristics} to enhance the richness and aptness of top-$k$ passages. 
    
    \paragraph{Default Ranking (def).}
    For a given SP pair, formulated as a natural language query, the dense retriever ranks 
    top-$d$ passages based on cosine similarity to the query vector.

    \paragraph{Entity Mention Frequency (num).} {\em re-ordering passages by frequency of named-entity mentions.} We detect mentions of entities (without disambiguation) of the proper type (usually person, place, or org) using spaCy and a hand-crafted dictionary of alias names for S and O, and paraphrases for P (incl. both nominal and verbal phrases). Top-$m$ passages with higher counts of mentions are prioritized, as they could potentially yield multiple O candidates.

    \paragraph{Amplification (amp).}
    {\em selecting support passages and re-ranking
    passages by pseudo-relevance feedback.}
    After extracting object lists from the initially selected passages, we assess the passage quality based on
    the no. 
    of objects they yield. The best passages, termed {\em support passages}, 
    are assumed to provide good cues about predicate P in surface form.  Following the rationale of pseudo-relevance feedback~\cite{DBLP:journals/ftir/Zhai08}, these support passages are fed back to the retriever for refined scoring, based on  embedding similarity between the candidate pool and the support passages.

    The {\em amp} technique works in two steps and iterates them as follows:
    \squishlist
    \item[1.] For each SP pair, we
    consider the previously generated O values and the best $s$ support passages: those from which the LLM could extract 
    the most objects (optionally weighted by extraction scores).
    \item[2.] All passages in the current pool are re-ranked by the retriever's scoring model based on combining the original query (about S and P) with the selected support passages.
    The now highest-ranking passages are then used for the next round of O extraction by prompting the LLM.
    \squishend
    Steps 1 and 2 are iterated in an alternating manner.
    For scoring by the embedding-based (dense) retriever, the refined query is a convex combination of the original query embedding and the sum of the top-$s$ support passages' vectors: 
    \vspace*{0.3cm}
    \\
    \vspace*{0.3cm}
    $~~~~~~~~~ \textbf{E}(Q') = \alpha{\textbf{E}(Q)} + (1 - \alpha){\sum_{i=1}^{s}{\textbf{E}(S_i)}}$\\
    \noindent with 
    embedding function 
    $\textbf{E( )}$ and hyper-parameter $\alpha$. 
    Algorithm~\ref{alg:amplification}
    gives pseudo-code for {\em amp}.
    
    \begin{algorithm*}[ht]
    {\small
    \caption{\small{Iterative Extraction with Pseudo-Relevance Feedback (amp Method).} 
    }
    \label{alg:amplification}
    \KwIn{  
        $\mathcal{C}$: candidate pool of retrieved passages;
        $Q$: retriever query in natural language with SP mentions;
        $k$: max no. of passages for prompting LLM;
        $b$: batch size;
        $s$: no. of support passages;
        $\alpha$: feedback weight for query reformulation;
        E: retriever's embedding function
    }  
    \KwOut{  
        List of object values $\mathcal{O}$  
    }  
    
    \KwInitialize{
    $\mathcal{O} \gets \emptyset$ ;
    $q \gets $ E($Q$) \Comment*[r]{\textcolor{blue}{\small{//embedding vector}}}
    }
    
    \For{ $i \gets 1$ to $\lceil k/b \rceil$}
    {  
        $\mathcal{K} \gets $ Retriever($\mathcal{C}$, $q$) \Comment*{\textcolor{blue}{\small{//ranking   $\mathcal{C}$ for top-k passages}}}
    
           ${p_b} \gets$ Batching($\mathcal{K}$) \Comment*{\textcolor{blue}{\small{//$b$ passages by def or by neo/sim (Section \ref{subsec:batching})}}}
        
        $\mathcal{O}_b \gets$ LLM(${p_b}$)  \Comment*{\textcolor{blue}{\small{//extracting objects by prompting LLM with passages $p_b$}}}
        $\mathcal{O} \gets \mathcal{O} \cup \mathcal{O}_b$ ;
        
        $\mathcal{S}_b \gets \emptyset$ ;\newline 
        \ForEach{passage $p \in p_b$} 
        {  
            {\If{$o \in \mathcal{O}_b$ appears in $p$}  
                {
                        $\mathcal{S}_b \gets \mathcal{S}_b \cup p$ \Comment*{\textcolor{blue}{\small{//finding support passages from passages $p_b$}}}
                }
            }
        }   
        $\mathcal{S} \gets$ Top($\mathcal{S}_b$) \Comment*[r]{\textcolor{blue}{\small{//selecting top-$s$ passages by \#objects}}}
        
        $q \gets \alpha \cdot q + (1 - \alpha) \cdot \frac{1}{s} \sum_{p \in \mathcal{S}} \textbf{E}(p)$ \Comment*[r]{\textcolor{blue}{\small{//convex combination to rerank $\mathcal{C}$}}}   
    }
    \Return $\mathcal{O}$ 
    }%
    \end{algorithm*}

    \subsection{Passage Batching}
    \label{subsec:batching}
    To feed passages into the LLM, the default approach combines successive ranks into small batches, as determined by the (re-)ranker. Alternatively, we can group or batch passages~\cite{DBLP:conf/icde/FanHFC00024} based on coherent story structures to aid the LLM in extracting O values. We devise two criteria for this purpose and batch:
    
    \squishlist
    
    \item {\bf Named Entity Overlap (neo):} passages with a large overlap in named entity mentions;
    
    \item {\bf Passage Similarity (sim):} passages whose textual embeddings have a high cosine similarity.
    
    \squishend
    For neo, we compute Jaccard similarity using min-hash sketches of named entities, while sim uses the OpenAI text-embedding model.
    Both techniques process a priority queue of passages as follows: for each rank $r$
    (starting with the highest rank, $r$=1), 
    find the $b$-1 most related passages from lower ranks $(r'$>$r)$ to form a batch and prompt the LLM. Mark all the batch passages as ``done'' and proceed with the next lower rank $(r'$>$r)$, which is not yet ``done''.

    \subsection{Prompt-based Object Generation}\label{sec:prompting}

    The retrieved top-$k$ passages 
    mention the subject in some form (e.g., first name, last name, or alias) and may contain other named entities. 
    We append the passages into the prompt context for retrieval-augmented list generation 
    \cite{DBLP:journals/csur/LiuYFJHN23,DBLP:journals/corr/abs-2312-10997,DBLP:journals/corr/abs-2303-18223}.
    As LLMs have limits on input context (and GPU memory demands
    increase with input length), we divide the top-$k$ passages (ranked by retriever scores) into batches of $b$ passages each (e.g., k=20, b=4 gives 5 batches). The O values generated from batch-wise processing are combined by their union for high recall.
    
    Prompts can optionally include a small set of demonstration examples for in-context inference. These examples explicitly mention SP appearing in books disjoint from the dataset, along with their complete O lists, aiding the instruction-tuned LLM in object list generation. We refer to this mode as {\bf few-shot} prompting, while the basic mode without examples is referred to as {\bf zero-shot}. Table~\ref{tab:prompt_template} shows an example for the few-shot prompt formulations.
    
    \begin{table}[ht]
        \centering
        \resizebox{\columnwidth}{!}{
        {\footnotesize
        \begin{tabular}{p{77mm}}
             \toprule
             \textit{System}:- \small{You are a knowledge base. Generate the complete list of names (Objects) who are parents, including step parents, of the specified person in the given book. List the names one after the other, separated by commas.}\\
             \midrule
             Few-Shot examples:\\[1ex]
             {Input:} \small{Book: A Promised Land, Subject: Barack Obama, Relation: parent}\\
             {Output}: \small{[Barack Obama Senior, Stanley Ann Dunham]}\\[1ex]
             {Input}: \small{Book: The Fellowship of the Ring, Subject: Frodo Baggins, Relation: parent}\\
             {Output}: \small{[Drogo Baggins, Primula Brandybuck]}\\        
            \midrule        
            \textit{User}:- Use the attached passages from the book.\\
            \small{Book:\{B\}, Subject:\{S\}, Relation:\{P\}, 
            Passages: \{T\}}\\
            \bottomrule
        \end{tabular}}
        }%
        \vspace{-0.5mm}
        \caption{Example of prompt template for Parent relation \\ (placeholders in curly brackets).}
        \label{tab:prompt_template}
        \vspace{-2mm}
    \end{table}

    In {\bf single-prompt} mode, the LLM uses only the best of these formulations (i.e., considered most natural by humans). In {\bf ensemble} mode, for each relation, we manually prepare five prompt templates for direct prompting, and five retriever query templates for L3X-RAG, and repeat all LLM-based extraction tasks with all templates. 
    The final O is the union of the O values generated across all runs.
    
    Of the four configurations (zero-single, zero-ensemble, few-single, few-ensemble), we report main results for the {\bf few-ensemble} setting, with the other configurations evaluated in ablation studies.

    \section{Precision-oriented Scrutinization}\label{sec:precision}

    To scrutinize the candidate objects O for a given SP and eliminate false positives, we devise several techniques.
    The key idea is to identify passages that clearly reflect SPO triples, and use these {\em support passages} to rank and prune O values, and also learn embeddings for the P predicates. 
    
    Figure~\ref{fig:methodology} (right side) gives a pictorial overview.
    From the first stage, each batch of passages yields an LLM-generated score for the output list of O values.
    The total score for this O (for a given SP) can be computed as a weighted occurrence frequency:
    $$\texttt{score}(O) = \sum_{\text{batch}_i} \exp{(\text{score}_{\text{LLM}}(L_i))} \times \textbf{I}_i(O)$$
    \noindent where $\textbf{I}_i(O)$ is an indicator variable set to 1 if O occurs in the output list $L_i$ for the i$^{th}$ batch of passages, and zero otherwise. $\text{score}_{\text{LLM}}$ is the log probability.
    This scoring serves as a simple baseline for pruning doubtful O values. 
    
    \subsection{Evidence Retrieval}
    
    While stage 1 needs to start the retrieval with S and P only, stage 2 has O candidates at its disposal.
    This allows us to search the entire book for textual snippets that explicitly indicate SPO triples. For each SPO candidate, we retrieve the top-$s$ passages, termed the {\em support passages} for SPO.
    Note that these are different from the support passages used by the {\em amp} method in stage 1,
    as we now retrieve from scratch from the entire book.
    
    To retrieve the support passages, we use the OpenAI text-embedding model to generate passage embeddings. These are compared against embeddings of the concatenated SPO strings, including SO alias names and paraphrases of P, using cosine similarity.
    
    \subsection{Classifiers}
    We devise several classifiers to scrutinize O values.
    
    \paragraph{Score-based Thresholding (thr).}
    As a baseline without support passages, the O candidates for a given SP are ranked using $\texttt{score}(O)$. We accept those that fall within the $t^{th}$ quantile (e.g., $t=0.8$) of the cumulative score distribution.
    
    \paragraph{Confidence Elicitation (conf).}
    We prompt the LLM again to assess its confidence in the generated O values.
    For each {\em SPO}, 
    top-$p$ support passages in their enriched form (with all named entities incl. S and O) are included into the LLM prompt for in-context inference:
    ``Given this information, is {\em SPO} a correct statement?''.
    The {\em conf} classifier accepts an O candidate if the LLM gives a ``yes'' reply.
    This approach differs from the passage-based extraction of the recall-oriented stage,
    as the support passages are retrieved individually for each O-candidate.
    
    \paragraph{Predicate-specific Classifier (pred).}
    The collection of support passages, for all SO with the same predicate P, can be utilized to learn an embedding for P cues, sort of a ``mini-LM'' for P.
    The intuition is that support passages with indicative phrases, such as
    ``life-or-death combat with'', ``deeply hates'' or
    ``I will destroy you'' (in direct speech), can collectively encode a better signal for P.
    To construct the classifier, we perform the following steps:
    
    \squishlist
    \item[1.] For each O, we retrieve top-$p$ support passages, and encode them into embedding vectors.
    \item[2.] We identify 
    the top-ranked O values with $\texttt{score}(O)$ above a threshold $\omega$.
    \item[3.] Using the top-ranked O, we combine the per-O passage vectors by a weighted sum, with $\texttt{score}(O)$ as weights, to obtain a single P-vector (classifier). 
    \item[4.] Each SO pair under scrutiny (O below the threshold $\omega$) is tested by comparing the vector of the top-$p$ support passages for this SPO candidate against the P-vector computed using steps 1 to 3. 
    \item[5.] The classifier accepts a low-ranked SO if the cosine similarity between the embeddings is above a threshold $\theta$.
    \squishend
    \noindent We construct a {\em pred} classifier for each SP pair, in a completely self-supervised manner.
    It has hyper-parameters $\omega$, $p$ and $\theta$, though; these are tuned via withheld train/dev data with SPO ground-truth, but without any supervised passage labels.

    \paragraph{Discriminative Classifier (dis).}
    Another way of harnessing the SPO support passages is to train a discriminative classifier, again in a self-supervised manner.
    We consider the ranked list of O values for a given SP and pick:
    \squishlist
    \item the top-$q$ high-scoring O candidates
    \item the bottom-$r$ low-scoring O candidates
    \squishend
    with $q$ and $r$ as hyper-parameters. For each top-$q$ and bottom-$r$ candidate O, we retrieve top-$p$ support passages, forming one passage pool for the high-scoring Os and another pool for the low-scoring Os.
    In each of these pools, the passages are cast into embeddings,
    and weighted averaged with $\texttt{score}(O)$ to form $SP_{high}$ and $SP_{low}$ vectors.
    
    Finally, each candidate O for a given SP is classified by whether its own support-passage vector is closer to the $SP_{high}$ or the $SP_{low}$ vector, in terms of cosine distance,
    leading to acceptance or rejection, respectively.

    \section{Experimental Setup}
\label{sec:experimental-setup}

\subsection{Datasets}\label{subsec:Datasets}
We make use of {\em fiction books} as a most representative, primary target for experimental studies. A second dataset, on {\em web contents with business-oriented relations}, exhibits different characteristics, and
adds diversity to the experiments. 
The results go into more depth and variety on the books data, and are shorter on the complementary web data.

\paragraph{Books Data.} The task of extracting long O lists from long texts is novel, with no suitable benchmark datasets available.
Therefore, we constructed a new dataset of books and corresponding ground-truth O lists associated with SP pairs.

We selected eleven popular novels and entire book series\footnote{A Song of Ice and Fire Series, Godfather Series, Harry Potter Series, Outlander Series, Little Women, Malibu Rising,  Pride and Prejudice, Steve Jobs, The Girl with the Dragon Tattoo, Wuthering Heights, The Void Trilogy}, enthusiastically discussed on community websites\footnote{\url{www.cliffsnotes.com}, \url{www.bookcompanion.com}, \url{www.fandom.com}}.
These fan communities feature extensive lists and infoboxes from which we derived SPO ground-truth with high confidence. As detailed in Sec~\ref{subsec:retriever}, the total no. of passages per book varies, ranging from ca. 10,000 passages in epic book series like A Song of Ice and Fire  to ca. 700 passages in shorter books like Malibu Rising.

Since entities often appear under multiple surface forms,
we manually constructed an entity name dictionary grouping alias names for each distinct entity. On a per-book basis, we ensured that certain first names, last names, or nicknames were uniquely identifiable.
For example, ``Daenerys'' is unique, whereas ``Targaryen'' is ambiguous. So for this entity, aliases include ``Daenerys'', ``Dany'', ``Daenerys Targaryen'', ``Daenerys Stormborn'', but not ``Targaryen''.
LLM outputs like ``Targaryen'' alone are thus counted as false.
This construction was aided by additional community sources\footnote{incl. \url{potterdb.com} for Harry Potter, \url{www.reddit.com/r/asoiaf} for Song of Ice and Fire, and others}.

This dataset comprises 764 distinct SP pairs for 8 predicates. 
In total, it covers ca. 5300 entities that appear under ca. 12,000 alias names. 
The S entities are typically prominent characters in the books, but they are associated with long O lists
mostly consisting of rarely mentioned long-tail entities.

\begin{table}[pt]
\resizebox{\columnwidth}{!}
{\begin{tabular}{|l||l|l|l|l|} \hline
\multirow{2}{*}{\textbf{Relation}} & \multirow{2}{*}{\textbf{Type}} & \multirow{2}{*}{\textbf{\#S}} & \multicolumn{2}{c|}{\textbf{\#O per S}} \\
\cline{4-5}
& & & \textbf{range} & $\mu$ ($\sigma$) \\ \hline\hline
parent & pers$\rightarrow$pers & 85 & 1--4 & 1.9 (0.6) \\ \hline
child & pers$\rightarrow$pers & 48 & 1--9 & 3.3 (2.4) \\ \hline
sibling & pers$\rightarrow$pers & 65 & 1--8 & 3.0 (1.8) \\ \hline\hline
family & pers$\rightarrow$pers & 81 & 1--47 & 12.1 (9.8) \\ \hline
friend & pers$\rightarrow$pers & 99 & 1--85 & 11.1 (16.5) \\ \hline
opponent & pers$\rightarrow$pers & 88 & 1--60 & 8.9 (11.2) \\ \hline
placeHasPer & loc$\rightarrow$pers & 189 & 1--92 & 6.7 (12.6) \\ \hline
hasMember & org$\rightarrow$pers & 109 & 1--142 & 11.6 (20.5) \\ \hline
\end{tabular}}
\caption{Books Dataset Statistics. \textbf{\#S} denotes the no. of unique subjects and \textbf{\#O} is the no. of objects per subject.}
\label{tab:BookDatasetStatistics}
\end{table}

\paragraph{Relation Difficulty.} The chosen 8 predicates has 3 {\em easier relations} with a relatively limited no. of O values (parent, child, and sibling) and 5 {\em harder relations} with potentially long O lists (family, friend, opponent, placeHasPerson---i.e., people being at a place, and hasMember---i.e., members of org. or events). Table~\ref{tab:BookDatasetStatistics} gives statistics for our dataset.

\paragraph{Web Data.}
To demonstrate the generalizability of L3X,
we constructed a second dataset with partly similar and partly complementary characteristics. The data is derived from the large common crawl of web pages (C4 corpus)~\cite{dodge-etal-2021-documenting},
and we aim to extract long object lists for three business/biography relations: CEOs of companies (incl. past ones), subsidiaries of companies, and organizations that a famous person is part of (e.g., companies, societies, charities, schools at different levels).
The dataset has 100 subjects for each P,
covering ca. 6400 entities appearing under ca. 24,000 alias names. 
Table \ref{tab:WebDataStatistics} gives per-relation statistics. 

\begin{table}[pt]
\resizebox{\columnwidth}{!}
{\begin{tabular}{|l||l|l|l|l|} \hline
\multirow{2}{*}{\textbf{Relation}} & \multirow{2}{*}{\textbf{Type}} & \multirow{2}{*}{\textbf{\#S}} & \multicolumn{2}{c|}{\textbf{\#O per S}} \\
\cline{4-5}
& & & \textbf{range} & $\mu$ ($\sigma$) \\ \hline\hline
hasCEO & org$\rightarrow$pers & 100 & 2-43 & 8.5 (5.2) \\ \hline
isMemberOf & pers$\rightarrow$org & 100 & 3-74 & 20.1 (14.7) \\ \hline
hasSubsidiary & org$\rightarrow$org & 100 & 2-308 & 71.8 (52.0) \\ \hline
\end{tabular}}
\caption{Web Dataset Statistics.
}
\label{tab:WebDataStatistics}
\end{table}

\subsection{System Configurations}
The presented methodology comes with many options for its different components, with the most important choices 
occurring for ranking, batching and scrutinization.
In our experiments, we focus on 
configurations for these three components,
labeling them accordingly, e.g., as {\em num/sim/thr} or {\em amp/neo/pred}. 
The {\em thr} classifier with default setting $t=0.8$ is abbreviated as {\em thr80}.

\paragraph{Hyper-Parameters.}
The L3X framework comes with
tunable hyper-parameters; Table~\ref{table:hyperparams} lists the default values for most experiments. We widely varied these settings, reporting
notable cases in Section~\ref{subsec:sensitivity} on sensitivity studies.

The best settings 
were identified using withheld train/dev data.
To this end, we split the entire dataset into three folds
(30:20:50),
through stratified sampling on books and SP pairs,
ensuring equal representation of varying O-list lengths in both folds. 
For each subject in train/dev, the 
complete O list is taken from the ground truth, 
to prevent information leakage into the test set. 

The best hyper-parameter values are determined through grid search,
maximizing the recall metric in Stage 1 and the R@P50 metric in Stage 2 (or alternatively AUC).
This is done for two modes: a single {\em global} value for a hyper-parameter, or {\em per-predicate} values, specific for each P.

\paragraph{Evaluation Metrics.} 
To obtain insights into the precision-recall trade-off and to assess the end-to-end goal of populating a knowledge base with high-quality SPO triples, the primary metric of interest is {\em Recall@Precision (R@P)}, focusing on the most interesting cases of R@P50 and R@P80 (50\% and 80\% correctness). Additionally, we report 
precision and recall, both before and after scrutinization. 
To further reflect on the inherent trade-off between precision and recall, we also compute the {\em area under the curve (AUC)} of the precision-recall curve.

All reported numbers are
{\em macro-averaged percentage} scores, computed as follows:
\squishlist
\item[1.] For each SP pair, we consider the resulting O list and compute the list's precision and recall.
\item[2.] We average the numbers over SP pairs for P.
\item[3.] We average the numbers over predicates P.
\squishend
Note that there is only a mild imbalance between different predicates and our calculation is based on the respective number of SP pairs rather than total O counts.
As a secondary metric we also compute {\em micro-averages} over all 
SP pairs and their O lists, regardless of the predicate, and report the findings.

\paragraph{Ground-Truth Variants.}
As detailed in Section~\ref{subsec:Datasets}, the ground-truth
O lists are derived from online sources with high quality control.
Inevitably, such lists are often incomplete, particularly for books with hundreds of minor characters.
Therefore, we also evaluate
by {\em pooling-based ground truth}, where the true positives within the union of all O values returned by all the methods is the complete ground truth.

\begin{table}[pt]
\footnotesize
\centering
{
\begin{tabular}{|l||c|} 
\hline
\textbf{Parameter}                                               & \textbf{Default} \\ 
\hline\hline
$l$: passage length (\#char)                                     & 1000 \\
passage overlap (\#char)                                         & 200  \\
$d$: \# retrieved passages                                       & 500   \\
$k$: top-$k$ passages                                            & 40  \\ 
\hline
$b$: \# passages per batch                                       & 2   \\ 
$m$: \# passages in \textit{num}                                 & 50   \\
$s$: \# support passages in \textit{amp}                         & 2 \\

$\alpha$: feedback weight for \textit{amp}                       & 0.7  \\
\hline
$t$: percentile retained for \textit{thr}                        & 0.8  \\ 
$p$: top-$p$ support passages for {\em conf}                     &  2   \\
$\omega$: cut-off score for O values in \textit{pred}            &  50  \\ 
$p$: top-$p$ passages for {\em pred} and {\em dis}               &  5   \\
$\theta$: acceptance bound for \textit{pred}                     & 0.85 \\ 
$q$: top-$q$ O candidates for {\em dis}                          &  50  \\
$r$: bottom-$r$ O candidates for {\em dis}                       &  50  \\
\hline
\end{tabular}}
\vspace*{-0.1cm}
\caption{Hyper-Parameters for L3X 
}
\label{table:hyperparams}
\vspace{-0.4cm}
\end{table}

\begin{table*}[pt]
\centering
{
\footnotesize
\begin{tabular}{|ll||ccc|ccccc|}
\hline
\multicolumn{2}{|c||}{\multirow{2}{*}{\textbf{Method}}} & \multicolumn{3}{c|}{\textbf{Stage 1 }} & \multicolumn{5}{c|}{\textbf{Stage 2 (thr80) }}\\
\cline{3-10}
& & P & R & AUC & P & R & AUC & R@P50 & R@P80 \\
\hline\hline
\multirow{4}{*}{\rotatebox[origin=c]{90}{LLM-only}}
& zero-single                 & \colorbox[gray]{0.9}{41.9} & 38.7          & 15.8          & \colorbox[gray]{0.9}{42.6} & 28.6 & 15.0 & 19.6 & 15.9 \\
& few-single                  & 38.5          & 43.4          & 18.2                       & 39.1 & 31.9 & 16.9 & 23.7 & 17.6 \\\cline{2-10}
& zero-ensemble      & 32.1 & \colorbox[gray]{0.9}{49.6} & 20.2 & 34.2 & \colorbox[gray]{0.9}{41.9} & 19.8 & 29.4 & 21.3 \\
& few-ensemble       & 34.1          & 47.7          & \colorbox[gray]{0.9}{20.5} & 37.0 & 39.5 & \colorbox[gray]{0.9}{20.5} & \colorbox[gray]{0.9}{31.4} & \colorbox[gray]{0.9}{21.9} \\
\hline\hline
\multirow{9}{*}{\rotatebox[origin=c]{90}{L3X (few-ensemble)}} 
& def            & 12.0          & 84.3          & 22.9          & 14.6 & 82.8 & 22.7 & 40.2 & 26.1 \\
\cline{2-10}
& +num           & 11.8          & \textbf{85.2} & 21.8          & 14.8 & \textbf{83.1} & 22.7 & 39.7 & 24.8 \\
& +amp           & 13.7          & 83.6          & \textbf{27.5} & 16.0 & 81.0 & \textbf{27.4} & \textbf{48.6} & \textbf{35.9} \\
\cline{2-10}
& +neo           & 12.6          & 83.8          & 22.4          & 15.5 & 81.6 & 23.0 & 39.1 & 25.1 \\
& +sim           & 12.5          & 84.2          & 22.2          & 15.3 & 82.5 & 23.2 & 39.4 & 23.5 \\
\cline{2-10}
& +num +neo      & 12.9          & 85.0          & 22.3          & 15.2 & 81.9 & 22.8 & 38.0 & 25.0 \\
& +amp +neo      & \textbf{14.1} & 83.4          & 27.1          & \textbf{16.7} & 81.6 & 26.9 & 47.7 & 35.4 \\
\cline{2-10}
& +num +sim      & 12.1          & 83.8          & 21.8          & 14.7 & 81.5 & 22.4 & 38.0 & 24.8 \\
& +amp +sim      & \textbf{14.1} & 83.4          & 27.1          & 16.0 & 80.5 & 26.3 & 47.0 & 33.8 \\
 \hline
\end{tabular}
}
\vspace*{-0.1cm}
\caption{Results (\%) for L3X Stage 1 Configurations on Books Data (ranking:\{num, amp\}; batching:\{neo, sim\}; top-$k$=40; default thresholding $(t=0.8)$ for Stage 2)}
\label{tab:book_stage1}
\vspace{-0.2cm}
\end{table*}

\section{Experimental Results}

We first present our findings on the books dataset,
as this is the primary representative of
our setting---see Subsections ~\ref{subsec:findingsBooksData} and ~\ref{subsec:drilldown}. Results for the web dataset are given in Subsection~\ref{subsec:resultsWebData}.
Due to space constraints,
we restrict the presented experiments to the most notable configurations, with additional discussions on hyper-parameter sensitivity 
and error cases in Subsections ~\ref{subsec:sensitivity} and ~\ref{subsec:errorAnalysis}. 
In all tables, the best value for a metric is in boldface,
and row-wise best values are shaded.

\subsection{Main Findings for Books Data}
\label{subsec:findingsBooksData}

\subsubsection{Stage 1: Recall-oriented Extractor.}
Table~\ref{tab:book_stage1} reports macro-averaged results for different configs of LLM-only 
and L3X-RAG generations.
The table shows both Stage 1 and Stage 2 results, but for Stage 2 all methods employ the
basic {\em thr} classifier with default 
hyper-parameter $t$=0.8.

The results clearly show the superiority of 
L3X over LLM-only, with major gains in recall (by Stage 1) and substantial improvements for R@P in Stage 2, reaching almost 50\% for R@P50.
We make the following key observations:

\noindent{\bf Baselines.} LLM-only methods perform poorly, even as a few-shot ensemble. Stage 1 recall saturates near 50\%, 
with AUC reaching ca. 20\%.

\noindent{\bf L3X at Stage 1.} 
All L3X configurations greatly improve recall, up to almost 85\%.
To calibrate these numbers, we also determined an oracle-based upper bound:
given all retrieved passages (top-$d$=500),
how many ground-truth objects do actually appear in at least one of these passages.
For the books dataset, this oracle suggests a recall of 88\%. So our results are very close to what can be spotted at all,
with a reasonable retriever budget.

The default configuration {\em def}
already achieves competitive performance, 
and the best option for recall is the {\em num}
method with entity-count-based re-ranking.
However, in terms of AUC, the method that shines most is the iterative {\em amp},
leading to an AUC value of 27.5\%.

\noindent{\bf Best L3X Config at Stage 2.}
The strong AUC of {\em amp} is a strong
starting point for the {\em thr}
classifier at Stage 2, where it achieves the best R@P values, with almost 50\% for R@P50---with a large margin over the second-best method.

\noindent{\bf Influence of Batching.}
When combined with batching via {\em neo}
or {\em sim}, L3X improves in precision, but loses recall, and eventually stays inferior to
{\em amp} alone.
While 
unexpected, it is not counter-intuitive:
{\em amp} operates iteratively, and judiciously picks its batch of passages in each round. So batching does help, but the big mileage already comes from the amplification of support passages.

\noindent{\bf Micro-averaging.}
With micro-averaging rather than macro-averages, 
the relative gains/losses across configs hardly change. The best results are
still with {\em amp}, reaching 84.7\% recall and 24.9\% AUC in Stage 1, and 44.7\% R@P50 and 31.6\% R@P80 in Stage 2.

\noindent{\bf Pooling-based Ground Truth.}

In this evaluation mode, stage-1 recall by {\em def} and {\em amp} increases by ca. 10 points, reaching 93\% and 92\%, respectively. 
This is intuitive, as pooled ground truth is a subset of the fully hand-crafted ground truth.
The numerical gains carry over to Stage 2: with {\em thr80}, {\em amp} goes up to 56.5\% R@P50 and 41.4\% R@P80. 

\noindent{\bf Influence of LLM Pre-Training.}
Most books in our data are well discussed in online media (incl. movie/TV adaptations). The LLM implicitly taps into this contents by its parametric memory.
To assess the influence of LLM pre-training, we evaluate LLM-only vs. L3X {\em amp} for one of our books, the Void Trilogy. This book series is barely covered on the Web;
we invested great effort for compiling its ground truth.
The results show that LLM-only fails completely on this case: 12\% recall with poor precision of ca. 5\%, whereas our RAG-based {\em amp} gets 82\% recall after Stage 1, and 34.1\% R@P50 and 38.3\% R@P80 
with default
{\em thr80} in Stage 2.

\subsubsection{Stage 2: Precision-oriented Scrutinizer.}

In the following, the presentation is restricted to a small subset of the
best performing stage-1 configurations, namely, 

{\em amp} and {\em amp/neo}, plus
the default L3X-RAG {\em def} for contrast.

Table~\ref{tab:recall_results} shows the stage-2 results with different %
classifiers for scrutinization.

We highlight the following key findings:

\noindent{\bf Best Configurations pred and dis.} The basic {\em thr} technique already works fairly well, 
especially with tuning of its hyper-parameter $t$. The more sophisticated classifiers {\em pred} and {\em dis}
still have an added benefit, improving the final R@P values by another 1\%, going up to 49.7\% for R@P50.

\noindent{\bf Hyper-parameter Tuning for pred.} The {\em pred} method has three hyper-parameters. Setting their values by global grid search with train/dev data leads to the best results, with $\omega$=20, $p$=5, and $\theta$=0.75.
As the various P exhibit different characteristics, we would expected even further gains with the per-predicate grid search.
Indeed, this led to rather different predicate-specifc values. For example, 
for Sibling, the best values are
$\omega$=10, $p$=2, $\theta$=0.9, whereas for
Friend we have $\omega$=50, $p$=1, $\theta$=0.55.
This makes sense, as Sibling lists are much shorter, and long Friend lists have much noisier support passages. So the setting is tighter %
for Sibling, and has more liberal $\theta$
for Friend but only 1 support passage per O to tame noise.
Nevertheless, the {\em pred $p$} and
{\em dis $p$} techniques did not achieve significant improvements over the globally tuned variants {\em pred $g$} and
{\em dis $g$}. We attribute this
result to the fact that the simpler configurations are already close to the best possible outputs (due to the difficulty of the task).

\noindent{\bf Confidence Elicitation from LLM.} {\em conf} performs poorly, as the LLM becomes rather conservative when fed with support passages about SPO candidates, rejecting too many valid entries.

\noindent{\bf Comparison to Relational IE.}
As Stage 2 processes full SPO triples, we can apply
relational IE
for classifying whether an SO pair satisfies P.
We considered two state-of-the-art methods,
GenIE~\cite{josifoski-etal-2022-genie} and DREEAM~\cite{ma-etal-2023-dreeam},
and tuned them on our predicates with the train/dev fold.
Still, both
performed very poorly, reaching less than 5\% recall and precision 10\% at best.
Clearly, our task is outside their comfort zone, as passages from fiction books are very different from Wikipedia paragraphs for which these methods were originally designed.
This underlines the uniqueness and challenging nature of the proposed long-lists-from-long-documents task.

\begin{table}[tp]
\centering
{
\footnotesize
\setlength{\tabcolsep}{2.2pt}
\begin{tabular}{|ll||ccccc|} 
\hline
\multicolumn{2}{|c||}{\textbf{L3X}} & P & AUC & R@P50 & R@P80 & R@P90 \\
\hline\hline
\multirow{7}{*}{\rotatebox[origin=c]{90}{def}} 
& thr $g$                 & 14.6 & 22.7 & 40.2 & 26.1 & 24.1 \\
& thr $p$                 & 16.5 & 22.1 & 39.5 & 25.9 & 23.9 \\
\cline{2-7}
& conf                    & 43.6 & 18.2 & 28.4 & 18.4 & 17.8 \\
\cline{2-7}
& pred $g$                & 18.9 & 23.4 & 41.3 & 26.6 & 24.8 \\
& pred $p$                & 21.5 & 22.9 & 40.7 & 26.5 & 24.7 \\
\cline{2-7}
& dis $g$                 & 20.2 & 23.1 & 41.2 & 26.6 & 24.8 \\
& dis $p$                 & 21.4 & 21.8 & 40.1 & 26.2 & 24.7 \\
\cline{2-7}
\hline\hline
\multirow{7}{*}{\rotatebox[origin=c]{90}{amp}} 
& thr $g$                 & 16.4 & 27.3 & 48.5 & 35.8 & 32.9 \\
& thr $p$                 & 17.5 & 27.2 & 48.5 & 35.8 & 32.9 \\
\cline{2-7}
& conf                    & \textbf{46.6} & 19.0 & 31.7 & 20.4 & 19.3 \\
\cline{2-7}
& pred $g$                & 20.4 & \textbf{28.1} & \textbf{49.7} & \textbf{36.5} & \textbf{33.3} \\
& pred $p$                & 23.5 & {28.0} & 48.7 & 36.2 & {33.0} \\
\cline{2-7}
& dis $g$                 & 21.5 & 27.9 & \textbf{49.7} & \textbf{36.5} & \textbf{33.3} \\
& dis $p$                 & 21.2 & 27.5 & 49.6 & 36.4 & \textbf{33.3} \\
\cline{2-7}
\hline\hline
\multirow{7}{*}{\rotatebox[origin=c]{90}{amp/neo}} 
& thr $g$                 & 16.4 & 26.9 & 47.7 & 35.4 & 33.1 \\
& thr $p$                 & 17.4 & 26.7 & 47.5 & 35.3 & 33.0 \\
\cline{2-7}
& conf                    & {45.4} & 18.7 & 30.7 & 20.1 & 19.6 \\
\cline{2-7}
& pred $g$                & 19.8 & 27.6 & 48.7 & 35.7 & 33.1 \\
& pred $p$                & 22.1 & 27.4 & 48.0 & 35.4 & 32.9 \\
\cline{2-7}
& dis $g$                 & 20.7 & 27.4 & 48.7 & 35.7 & 33.1 \\
& dis $p$                 & 20.9 & 26.2 & 48.0 & 35.4 & 32.8 \\
\cline{2-7}
\hline
\end{tabular}
}
\vspace*{-0.1cm}
\caption{Results (\%) for L3X Stage 2 Configurations (classifiers: \{thr, conf, pred, dis\}, 
$g$ refers to global grid search and $p$ is per-predicate grid search).}
\label{tab:recall_results}
\end{table}

\begin{table*}[tbh]
\centering
{
\footnotesize
\setlength{\tabcolsep}{4pt}
{\begin{tabular}{|l||ccc||ccc|ccc|ccc|} \hline
{\multirow{2}{*}{\textbf{L3X}}} & \multicolumn{3}{c||}{\textbf{stage 1: recall}} & \multicolumn{3}{c|}{\textbf{def/pred-$g$}} & \multicolumn{3}{c|}{\textbf{amp/pred-$g$}} & \multicolumn{3}{c|}{\textbf{amp/neo/pred-$g$}}\\ 
\cline{2-13}
            & def & amp & amp/neo & AUC &  R@P50 & R@P80 & AUC & R@P50 & R@P80 & AUC & R@P50 & R@P80 \\ 
            \hline\hline
parent      & 75.6 & \colorbox[gray]{0.9}{76.2} & 73.8 & 27.3 & 57.7 & 47.0 & 27.9 & 61.3 & 52.4 & \colorbox[gray]{0.9}{27.9} & \colorbox[gray]{0.9}{61.3} & \colorbox[gray]{0.9}{53.6} \\
children    & \colorbox[gray]{0.9}{86.5} & 82.5 & 84.6 & 28.1 & 60.9 & 43.8 & \colorbox[gray]{0.9}{36.5} & \colorbox[gray]{0.9}{72.3} & \colorbox[gray]{0.9}{61.0} & 34.4 & 66.1 & 55.2 \\
sibling     & 87.2 & 86.2 & \colorbox[gray]{0.9}{89.3} & 38.0 & 65.4 & 50.4 & \colorbox[gray]{0.9}{47.7} & \colorbox[gray]{0.9}{79.4} & \colorbox[gray]{0.9}{67.7} & 46.3 & 77.9 & 66.0 \\
\hline
\textbf{avg. Easy P} & \colorbox[gray]{0.9}{83.1} & 81.6 & 82.6 & 31.1 & 61.3 & 47.1 & \colorbox[gray]{0.9}{37.3} & \colorbox[gray]{0.9}{71.0} & \colorbox[gray]{0.9}{60.3} & 36.2 & 68.4 & 58.3 \\
\hline\hline
family      & 79.8 & \colorbox[gray]{0.9}{79.8} & 78.5 & 25.2 & 34.0 & 15.0 & \colorbox[gray]{0.9}{33.1} & 44.8 & \colorbox[gray]{0.9}{33.2} & 32.9 & \colorbox[gray]{0.9}{46.2} & 32.1 \\
friend      & 85.4 & \colorbox[gray]{0.9}{85.5} & 85.2 & 19.7 & 27.1 & 13.1 & \colorbox[gray]{0.9}{23.8} & 35.5 & 17.0 & 23.2 & \colorbox[gray]{0.9}{35.5} & \colorbox[gray]{0.9}{18.6} \\
opponent    & 80.8 & \colorbox[gray]{0.9}{81.1} & 80.1 & 17.6 & 29.3 & 14.5 & \colorbox[gray]{0.9}{18.9} & \colorbox[gray]{0.9}{32.4} & \colorbox[gray]{0.9}{14.6} & 18.8 & 30.9 & 14.2 \\
hasMember   & \colorbox[gray]{0.9}{89.0} & 86.6 & 86.1 & 16.5 & 25.7 & 14.0 & 20.7 & 32.5 & 20.9 & \colorbox[gray]{0.9}{21.1} & \colorbox[gray]{0.9}{36.5} & \colorbox[gray]{0.9}{20.9} \\
placeHasPer & 89.8 & \colorbox[gray]{0.9}{90.7} & 89.4 & 14.7 & 30.3 & 15.3 & \colorbox[gray]{0.9}{16.5} & \colorbox[gray]{0.9}{39.6} & \colorbox[gray]{0.9}{25.0} & 16.1 & 35.6 & 24.8 \\
\hline
\textbf{avg. Hard P} & \colorbox[gray]{0.9}{85.0} & 84.7 & 83.9 & 18.7 & 29.3 & 14.4 & \colorbox[gray]{0.9}{22.6} & \colorbox[gray]{0.9}{37.0} & \colorbox[gray]{0.9}{22.1} & 22.4 & 36.9 & 22.1 \\
\hline\hline
\textbf{avg. All P}  & \textbf{84.3} & 83.6 & 83.4 & 23.4 & 41.3 & 26.6 & \textbf{28.1} & \textbf{49.7} & \textbf{36.5} & 27.6 & 48.7 & 35.7 \\
\hline
\end{tabular}}
}%
\vspace*{-0.2cm}
\caption{Drill-Down Results by Predicate for Books Dataset
}
\label{tab:book_drilldownpred}
\end{table*}

\subsection{Analysis by Drill-Down}
\label{subsec:drilldown}

The reported results are macro-averaged over all relations. However, some relations P are easier to deal with than others (see Sec~\ref{sec:experimental-setup}). Table~\ref{tab:book_drilldownpred} shows results with drill-down by P,
comparing the best configs. after Stage 1 and Stage 2, respectively.

For stage-1 recall, we achieve similar performance across predicates, all between ca. 75\% and 90\% recall.
For stage-2 R@P results, 
we clearly see the big gap between "easy" relations with crisp lists of a few objects, and "hard" relations
with long lists and more loosely defined cues in the text passages. Not surprisingly, the Opponent relation is the hardest case, where even our best method reaches only ca. 30\% for R@P50.
This calls for more research on this challenging task.

\begin{table}[t]
\centering
\footnotesize
\setlength{\tabcolsep}{2.7pt}
\begin{tabular}{|ll||ccc|ccccc|}
\hline
\multicolumn{2}{|c||}{\multirow{2}{*}{\textbf{L3X}}} & \multicolumn{3}{c|}{\textbf{Stage 1}} & \multicolumn{5}{c|}{\textbf{Stage 2 (thr80)}}\\
\cline{3-10}
& & P & R & AUC & P & R & AUC & R@50 & R@80 \\
\hline\hline
\multirow{2}{*}{\rotatebox[origin=c]{90}{def}}
& EH & 18.0 & 90.3 & 26.3 & 22.2 & 89.3 & 26.4   & 65.9 & 46.1 \\
& ET & 16.7 & 79.3 & 15.3 & 19.9 & 74.9 & 16.8   & 35.7 & 26.0 \\
\cline{2-10}
\multirow{2}{*}{\rotatebox[origin=c]{90}{amp}}
& EH & 23.4	& 87.2 & 32.1 & 26.1 & 83.0 & 31.5   & 74.0 & 55.1 \\
& ET & 17.3	& 75.9 & 18.9 & 21.1 & 73.1 & 20.2   & 45.5 & 35.0 \\
\hline\hline
\multirow{2}{*}{\rotatebox[origin=c]{90}{def}}
& HH & 2.4  & 92.1 & 17.3 & 3.0 & 91.6 & 17.4   & 32.4 & 16.6  \\
& HT & 3.0  & 75.3 & 8.5  & 3.6 & 73.7 & 8.3   & 8.0 & 3.1    \\
\cline{2-10}
\multirow{2}{*}{\rotatebox[origin=c]{90}{amp}}
& HH & 2.5  & 91.9 & 20.6 & 3.2 & 91.1 & 20.4   & 40.5 & 24.7 \\
& HT & 3.1  & 74.9 & 10.4 & 3.5 & 72.1 & 10.4   & 13.1 & 4.1 \\
\hline
\end{tabular}
\caption{Results on Head-vs-Tail. EH: easy P+head, ET: easy P+tail, HH: hard P+head and HT: head P+tail.}
\label{tab:head_tail}
\end{table}

\paragraph{Entity Popularity.} 
For drill-down analysis, we partition the test set ground-truth O entities into {\em head} and {\em tail} groups. Based on the frequency of each unique O in a book, we define the head as entities appearing above the 75th percentile and the tail as those below it.
Table \ref{tab:head_tail} shows these results on four combinations (easy P, head), (easy P, tail), (hard P, head), and (hard P, tail). We observe that 
{\em amp} still outperforms {\em def} for all four cases.
However, for the most challenging case of hard P and tail O,
the absolute numbers degrade strongly.
This highlights the complexity of the task and the need for further research.

\subsection{Findings for Web Data}
\label{subsec:resultsWebData}

Table \ref{tab:web_stage1} shows results for the Web dataset, with different stage-1 configurations, and default {\em thr80} for scrutinization.
By and large, the results reconfirm our key findings with the Books data:
LLM-only methods are far inferior;
all L3X-RAG methods boost recall;
the {\em amp} has the best AUC after Stage 1 and is the overall winner for R@P after Stage 2.
The oracle-based upper bound here is 65\% for recall from top-$d$=500 passages.

A significant difference to the Books data, however, is that all absolute values are substantially lower here (incl. the oracle). For example, the best R@P50 values (by {\em amp}) are
around 30\%, compared to almost 50\% for the books experiment.
The explanation clearly is that the dataset itself is even more challenging: the ground-truth lists of objects are even longer, with even more long-tail entities, and they are spread across a very large number of web pages---a situation as if
all pages (about the same S) were concatenated into a very long, highly incoherent document).

Table \ref{tab:web_stage2} compares stage-2 classifiers, along with drill-down by the three predicates. The trends align with those observed in the books dataset, with {\em amp/pred g} achieving the highest scores. While performance is strong on the relatively easier CEO relation (ca. 65\% R@P50), it struggles on the highly challenging hasSubsidiary relation.

\begin{table}[t]
\centering
{
\footnotesize
\setlength{\tabcolsep}{1pt}
\begin{tabular}{|l||ccc|ccccc|}
\hline
\multirow{2}{*}{\textbf{Method}} & \multicolumn{3}{c|}{\textbf{Stage 1}} & \multicolumn{5}{c|}{\textbf{Stage 2 (thr80)}}\\
\cline{2-9}
& P & R & AUC & P & R & AUC & R@50 & R@80 \\
\hline\hline
zero-ens    & 25.0   & \colorbox[gray]{0.9}{43.5} & \colorbox[gray]{0.9}{19.0}   & 28.1 & \colorbox[gray]{0.9}{39.7} & \colorbox[gray]{0.9}{17.3} & \colorbox[gray]{0.9}{25.9} & \colorbox[gray]{0.9}{15.5} \\
few-ens      & 28.3 & 41.9 & 18.5 & \colorbox[gray]{0.9}{31.8} & 37.8 & 16.9 & 25.2 & 15.2 \\
\hline\hline
def       & 4.1  & \textbf{70.6} & 18.0   & 4.9  & \textbf{67.9} & 17.8 & 21.5 & 7.2  \\
num       & 4.4  & 70.3 & 17.9 & 5.2  & 67.9 & 17.4 & 20.8 & 8.3  \\
amp       & 13.4 & 60.5 & \textbf{23.5} & 15.7 & 57.8 & \textbf{23.3} & \textbf{31.9} & \textbf{18.3} \\
amp/neo   & 18.5 & 51.8 & 20.5 & \textbf{21.0} & 48.8 & 19.5 & 29.3 & 13.6 \\
 \hline
\end{tabular}
}\caption{Results on Web Data.
Stage-1 with ranking:\{num, amp\}; batching:\{neo\}; $k$=40; Stage-2: thr80.}
\label{tab:web_stage1}
\vspace{-0.3cm}
\end{table}

\begin{table*}[ht]
\centering
{\footnotesize
\setlength{\tabcolsep}{3pt}
{\begin{tabular}{|l||ccc|ccc|ccc|ccc|} \hline
{\multirow{2}{*}{\textbf{}}} & \multicolumn{3}{c|}{\textbf{def/thr80}} & \multicolumn{3}{c|}{\textbf{amp/thr80}} & \multicolumn{3}{c|}{\textbf{amp/pred($g$)}} & \multicolumn{3}{c|}{\textbf{amp/neo/pred($g$)}}\\ 
\cline{2-13}
& AUC & R@P50 & R@P80 & AUC & R@P50 & R@P80 & AUC & R@P50 & R@P80 & AUC & R@P50 & R@P80 \\ \hline\hline

hasCEO          & 33.5 & 51.0 & 19.5   & 41.1 & 62.0  & 42.6 &  44.7 & 64.7 & 44.4 & 42.9 &	63.2 &	42.9 \\
isMemberOf      & 12.1 & 10.0 & 1.2    & 15.9 & 21.8  & 8.3  & 16.4 & 23.2 & 9.1 & 15.6 & 22.2 & 9.9 \\
hasSubsidiary   & 7.8  & 3.4  & 0.9    & 12.8 & 12.0  & 4.1  & 13.1 & 13.0 & 5.9 & 12.9 & 12.3 & 4.0 \\
\hline\hline
{\bf macro-avg.}  & 17.8 & 21.5 & 7.2  & 23.3 & 31.9 & 18.3  & \textbf{24.7} & \textbf{33.6} & \textbf{19.8} & 23.8  & 32.6 & 18.9 \\\hline
\end{tabular}}
}%
\vspace{-0.2cm}
\caption{Results (\%) for Web Data with Drill-Down Results by Predicate. 
}
\label{tab:web_stage2}
\end{table*}

\begin{figure}[tp]
  \centering
  \includegraphics[width=\columnwidth, 
                   trim={0cm 0cm 0cm 0cm},
                    clip=true]{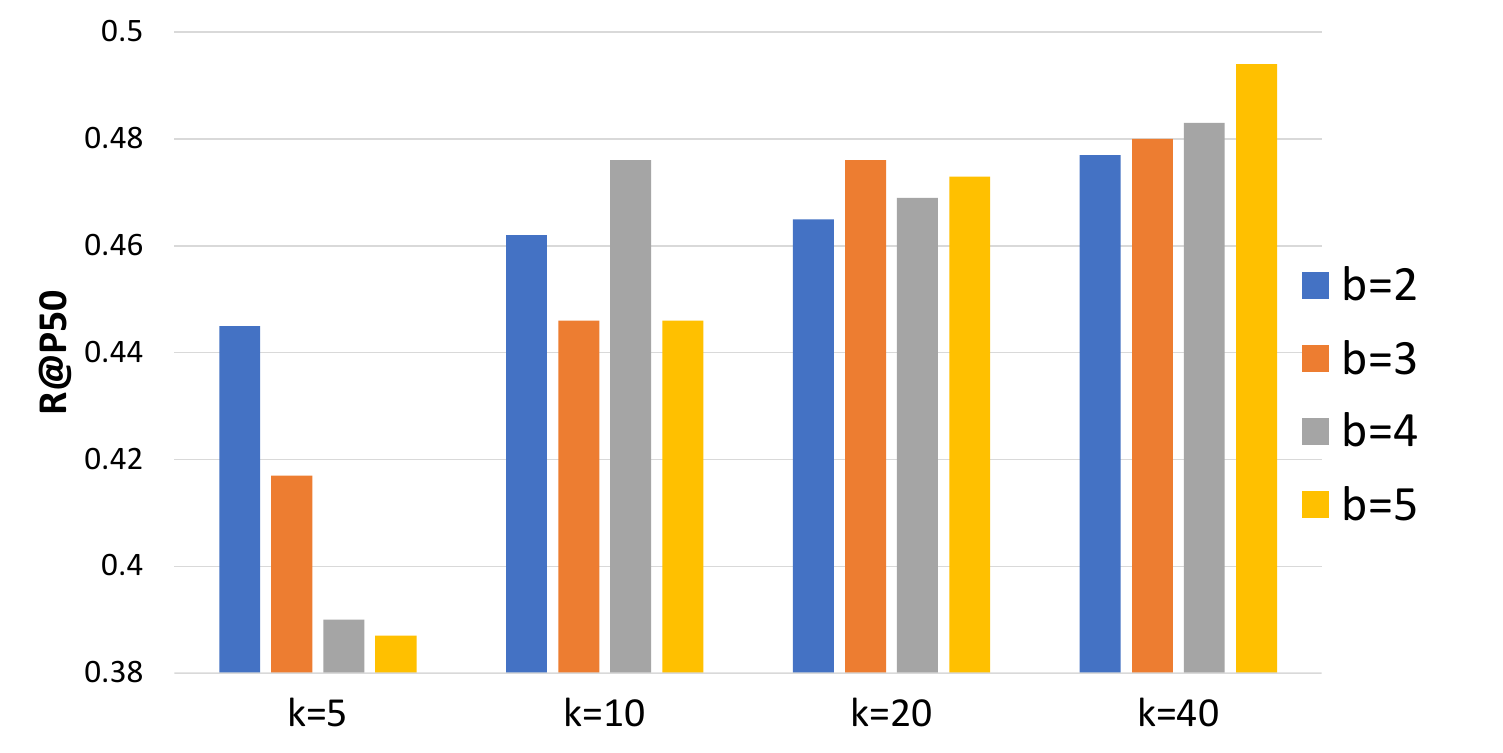}
  \vspace{-0.8cm}
  \caption{Varying hyper-parameters $k,b$ for {\em amp/neo}.}
  \label{fig:sensitivity_curve}
\end{figure}

\subsection{Sensitivity of Hyper-Parameter Settings}
\label{subsec:sensitivity}

We performed extensive experiments on varying hyper-parameter settings. We report on the sensitivity of the two most important Stage 1 choices: $k$ (no. top-$k$ passages) and $b$ (no. passages per batch). Figure \ref{fig:sensitivity_curve} shows {\em R@P50} numbers for {\em amp/neo} by varying $k$ from 5 to 40 and $b$ from 2 to 5.

We observe that increasing $k$ is very useful, but has diminishing returns beyond a certain value. Note that larger $k$ also increase the cost for invoking the LLM more often.
For small $k$, there is no benefit from larger batches, but this
changes for large $k$, where bigger batches keep improving recall.
Note that this also increases LLM costs, as we pass more tokens into
the inference.

All L3X-RAG configs use five reformulations for retriever queries. Table~\ref{tab:ablation} presents the change in results with fewer query variants. Both recall and R@P values drop with less queries, showing the vital role of diversified formulations.

\begin{table}[pt]
\centering
\footnotesize
\setlength{\tabcolsep}{2pt}
\begin{tabular}{|l||ccc|ccccc|}
\hline
\textbf{L3X} & \multicolumn{3}{c|}{\textbf{Stage 1}} & \multicolumn{5}{c|}{\textbf{Stage 2 (thr80)}}\\
\cline{2-9}
\textbf{{\em amp}} & P & R & AUC & P & R & AUC & R@50 & R@80 \\
\hline\hline
\#q=5         & 13.7 & 83.6 & 27.5 & 16.0 & 81.0 & 27.4 & 48.6  & 35.9  \\
\#q=3      & 15.2 & 81.1 & 27.2 & 18.1 & 78.5 & 27.5 & 48.2  & 35.4  \\
\#q=1       & 20.5 & 76.2 & 26.1 & 23.4 & 71.1 & 26.5 & 44.0  & 31.5  \\
\hline
\end{tabular}
\vspace{-0.2cm}
\caption{Varying no. queries for retriever with {\em amp}.}
\label{tab:ablation}
\vspace{-0.3cm}
\end{table}

\subsection{Error Discussion}
\label{subsec:errorAnalysis}

We observed a variety of typical error cases, and discuss three of the most notable ones.

\noindent{\bf Hallucinations.}
The LLM calls often return extremely long lists of objects, including names that do not occur in the respective books at all.
In RAG mode, 
the LLM does not necessarily restrict its outputs to names occurring in the input passages.
These are {\em unfaithful} generations, but for recall, our main target, producing names from parametric memory is an advantage.
To quantify the issue of hallucinated O values, we compute 
the no. of generated objects that do not appear in the respective book:\\
$~~~~~~~~~|\cup_{SP} \{\text{generated O for SP} ~|~ O \notin \text{book}\}|$\\
normalized by the total no. of generated O values.
We observed the following hallucination rates after Stage 1:
LLM-only: 55.3\%, def: 51.7\%, amp: 40.7\%, amp/neo: 38.1\%. 
Erroneous objects include made-up names and non-entity phrases, such as ``X's sister'', where the LLM appends a phrase related to the predicate instead of an O name. 
This underlines that stage-2 scrutinizing is essential.

\noindent{\bf Confusing Predicates.}
Another common case is that the LLM generates valid O values that are not in the proper relation P with subject S.
The most interesting situation here is when that incorrect O is in relation with S for another predicate Q ($\ne$ P)
(e.g., Dumbledore appearing among Harry Potter's parents instead of being a friend).

To quantify, we compute a \#P$\times$\#P
confusion matrix, with counts of generated O for P when ground truth is Q.
For our best method, {\em amp/pred(g)}, we observed a ratio of ca. 60:30:10 of accepted true positives (TP), predicate-confused TPs, and accepted false positives (FP).
This suggests that merely extracting the right SO pairs is not the problem (only 10\% completely FPs), but getting the predicate correct is the big issue here.
The most salient predicate pairs of
confusion are (Friend,Family) and (Friend,Opponent).
This may be surprising on first glance, but it is the
sophistication and subtlety of fictional literature that 
makes this a daunting case.

\noindent{\bf Missing True Positives in the Low Ranks.} Majority of TPs are in the higher ranks. These are followed by a long tail, with mostly FPs but sprinkled with TPs at lower ranks. 

To assess how well Stage 2 recovers {\em low-ranked} TPs, we identify the ranking cut-off for R@P50 and count the missing TPs below this threshold---i.e., those misclassified as false negatives.
Even with our best methods, ca. 16\% of all the ground-truth O values fall into the category of low-rank missing TPs.

\section{Conclusion}
We introduced the task of extracting long lists of objects %
from long documents,
and developed the L3X methodology, comprising LLM prompting, retrieval augmentation, passage ranking and batching, and classifiers to scrutinize candidates and prune false positives.
Extensive experiments with a range of L3X configurations over two datasets provide key insights.
First, L3X greatly outperforms LLM-only extraction in recall and R@P.
Second, one of our methods, {\em amp/pred} with pseudo-relevance feedback and a classifier with tuned hyper-parameters, achieves
remarkable performance of ca. 85\% recall and ca. 37\% R@P80
on full-length books.
However, drill-down analyses by predicate and head-vs-tail entities
show that for the hardest cases, 
there is substantial room for improvement.
Third, this underlines the biggest challenge of our 
task:
passages scattered across long books 
give cues 
to a smart human, but are still very hard to pinpoint and extract for AI systems (incl. LLMs).

\section{Limitations}

This work is based on Llama3.1-70B. 
We also ran Stage 1 studies using other LLMs.
With a 10x smaller Llama-8B model, recall fell below 50\% (compared to ca. 80\% with the 70B model).
With GPT-3.5, preliminary experiments showed results similar to Llama-70B. 
Still, a broader comparison of different LLMs for our task would be desirable.

In the absence of suitable datasets, we constructed a new benchmark resource---limited in scale, though. Expanding the dataset would be useful
, however, scaling up the benchmark for books poses a major challenge. On one hand, for popular books, data for ground truth construction is readily available, but these books are likely well captured in the model's parametric memory
from pre-training.
On the other hand, we could consider books in the long tail of popularity or brand-new books, but this incurs high cost to obtain ground truth, 
requiring end-to-end reading and great diligence for annotations.

\clearpage
\newpage
\bibliographystyle{acl_natbib}
\bibliography{custom,anthology}

\end{document}